\documentclass[runningheads]{llncs}
\usepackage[T1]{fontenc}
\usepackage{mathtools} 
\usepackage{graphicx,verbatim}
\usepackage{amsfonts,bbold}
\usepackage{booktabs}
\usepackage{amssymb}  
\usepackage{pifont}   
\usepackage{hyperref}
\usepackage{ulem}

\usepackage{color}

\begin{document}
\title{
Endo-TTAP: Robust Endoscopic Tissue Tracking via Multi-Facet Guided Attention and Hybrid Flow-point Supervision
}
\titlerunning{Track Any Points of Endoscopic Tissue}
%

\author{Rulin Zhou\inst{1,3~\star} 
\and Wenlong He\inst{1~\star} 
\and An Wang\inst{2} 
\thanks{Rulin Zhou, Wenlong He and An Wang are co-first authors.} 
\and Qiqi Yao\inst{1}
\and Haijun Hu\inst{4}
\and Jiankun Wang\inst{5}
\and Xi Zhang\inst{1}
\and Hongliang Ren\inst{2,3}\thanks{Corresponding author}}
\authorrunning{R. Zhou et al.}
\institute{College of Mechatronics and Engineering, Shenzhen University, Shenzhen, China
\and Dept. of Electronic Engineering, Shun Hing Institute of Advanced Engineering (SHIAE), The Chinese University of Hong Kong, Hong Kong SAR, China
\and CUHK Shenzhen Research Institute, Shenzhen, China
\and Division of Gastrointestinal Surgery, Department of General Surgery, Shenzhen People’s Hospital , Shenzhen, China
\and College of Engineering Southern University of Science and Technology (SUSTech), Shenzhen, China\\
\email{zhourulin2020@email.szu.edu.cn, hewenlong2023@email.szu.edu.cn, wa09@link.cuhk.edu.hk, yaoqiqi2022@email.szu.edu.cn, szphhhj@outlook.com, wangjk@sustech.edu.cn, hlren@ee.cuhk.edu.hk, zh0005xi@szu.edu.cn}
}
    
\maketitle              
\begin{abstract} 

Accurate tissue point tracking in endoscopic videos is critical for robotic-assisted surgical navigation and scene understanding, but remains challenging due to complex deformations, instrument occlusion, and the scarcity of dense trajectory annotations. Existing methods struggle with long-term tracking under these conditions due to limited feature utilization and annotation dependence.
We present \textbf{Endo-TTAP}, a novel framework addressing these challenges through: (1) A \textbf{Multi-Facet Guided Attention} (MFGA) module that synergizes multi-scale flow dynamics, DINOv2 semantic embeddings, and explicit motion patterns to jointly predict point positions with uncertainty and occlusion awareness; (2) A two-stage curriculum learning strategy employing an \textbf{Auxiliary Curriculum Adapter} (ACA) for progressive initialization and hybrid supervision. Stage I utilizes synthetic data with optical flow ground truth for uncertainty-occlusion regularization, while Stage II combines unsupervised flow consistency and semi-supervised learning with refined pseudo-labels from off-the-shelf trackers.
Extensive validation on two MICCAI Challenge datasets and our collected dataset demonstrates that Endo-TTAP achieves state-of-the-art performance in tissue point tracking, particularly in scenarios characterized by complex endoscopic conditions. The source code and dataset will be available at~\url{https://anonymous.4open.science/r/Endo-TTAP-36E5}.


\keywords{Endoscopic Tissue Tracking \and Hybrid Supervision \and Multi-Faceted Guided Attention}


\end{abstract}
\section{Introduction}

\begin{figure}[t]
\centering
\includegraphics[width=0.95\textwidth]{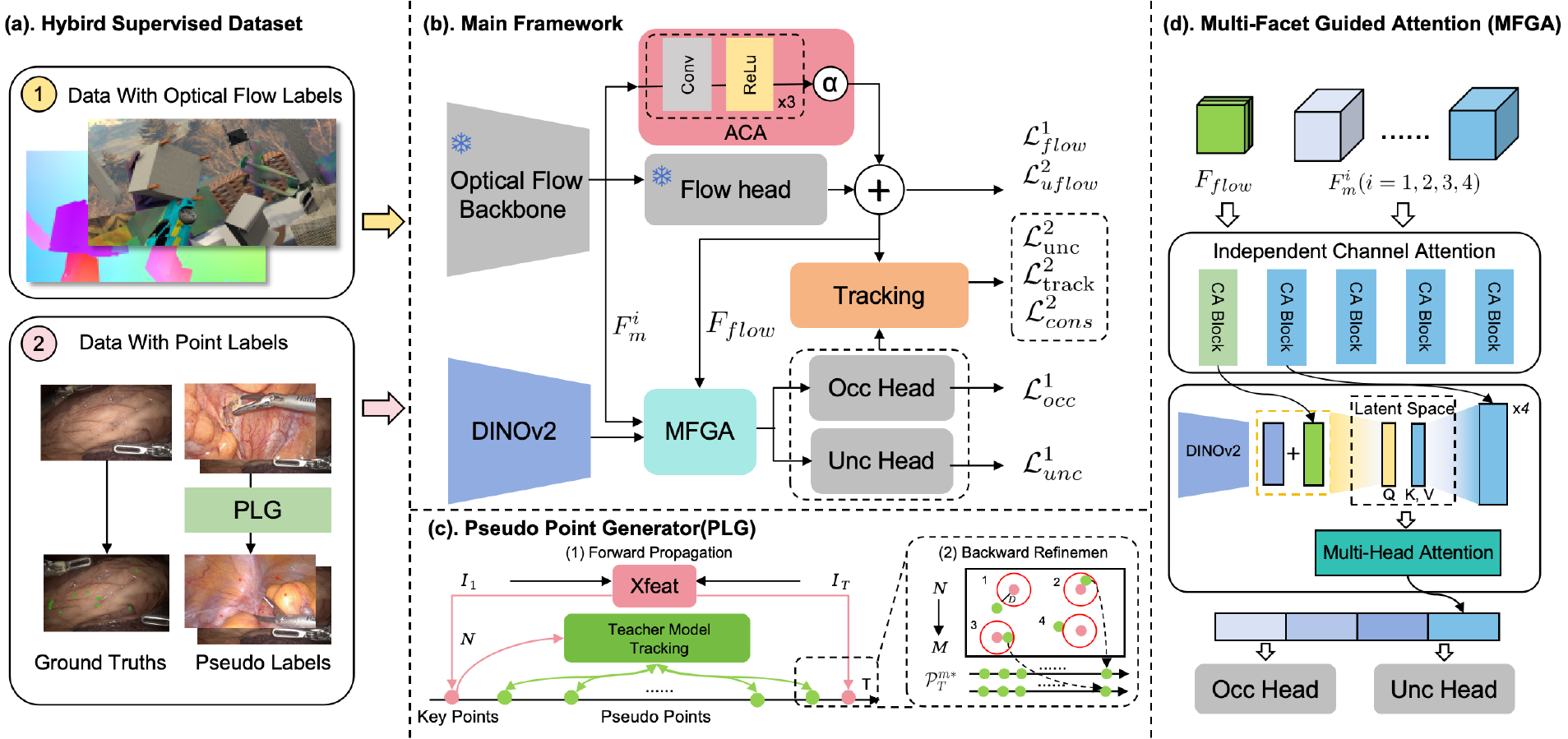}
\caption{Overview of our Endo-TTAP framework. The Auxiliary Curriculum Adapter (ACA) module facilitates progressive initialization and finetuning of the \textit{Uncertainty Head} and the \textit{Occlusion Head}. Hybrid datasets are utilized in the two-stage training of a robust tissue point tracking model with Multi-Facet Guided Attention (MFGA).} \label{fig:main}
\end{figure}

The rapid advancement of surgical robotics, exemplified by systems like the da Vinci Xi, has enabled minimally invasive procedures with reduced trauma and high precision~\cite{reddy2023advancements,rivero2023robotic}. Tissue point tracking, as a critical low-level task for surgical robotic intelligence, plays a vital role in applications such as surgical navigation and scene understanding~\cite{sheetz2020trends,long2023human}. However, endoscopic environments present unique tracking challenges, including complex tissue dynamics from respiration and instrument manipulation, as well as environmental artifacts like specular reflections, occlusions, and surgical smoke~\cite{fu2021future}. These factors collectively degrade the performance of conventional tracking algorithms~\cite{marmol2019dense,song2018mis}, which exhibit large mean endpoint errors in multi-instrument scenarios.

Recent advances in arbitrary point tracking follow two primary paradigms. Vision-based methods, such as PIPs~\cite{harley2022particle} and CoTracker~\cite{karaev2024cotracker,karaev2024cotracker3}, employ deep neural networks to establish dense correspondences but struggle with the homogeneous textures characteristic of endoscopic tissue. Optical flow approaches like RAFT~\cite{teed2020raft}, OmniMotion~\cite{wang2023tracking}, MFT~\cite{neoral2024mft}, and MFTIQ~\cite{serych2024mftiq}exploit motion continuity but require dense annotations unavailable in surgical settings. While datasets such as SurgT~\cite{allan2021stereo} and STIR~\cite{schmidt2024surgical} have spurred progress, existing solutions, including graph neural networks-based depth and flow refinement~\cite{schmidt2023sendd}, adaptive-template matching~\cite{guo2024ada}, and those from their respective challenges, remain inadequate for clinical deployment. Three fundamental limitations persist: (1) reliance on sparse annotations, (2) domain mismatch between natural video priors and surgical tissue biomechanics, and (3) lack of explicit uncertainty modeling for error propagation in occluded regions.

\begin{figure}[t]
\centering
\includegraphics[width=0.9\textwidth]{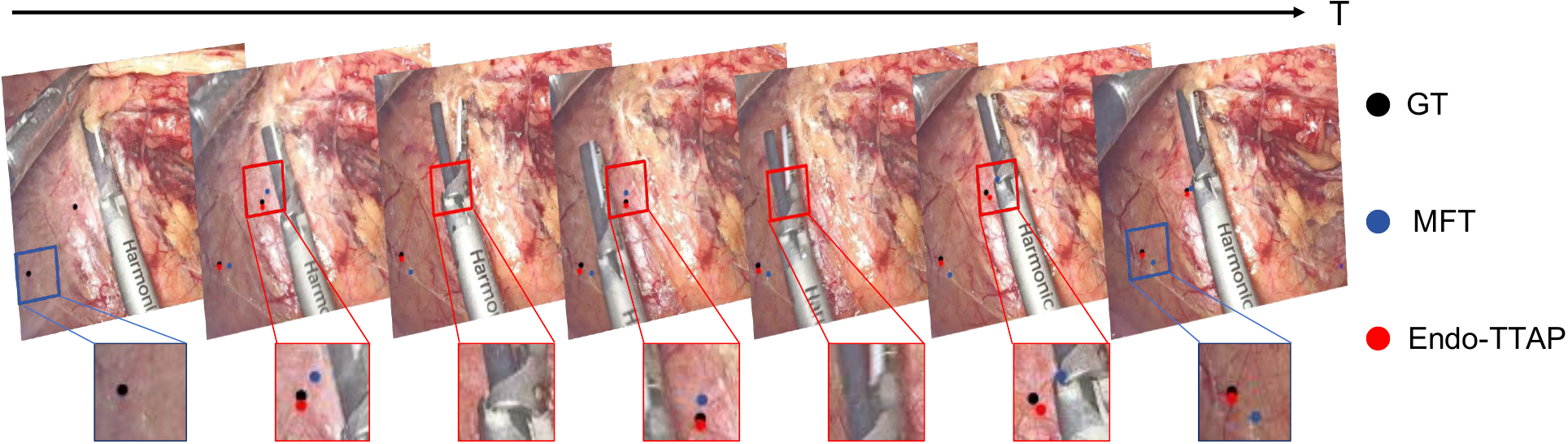}
\caption{Tissue point tracking comparison of our method (red point) with GT (black point) and MFT~\cite{neoral2024mft} (blue point). Our method exhibits superior tracking results in handling long videos (blue box) and instrument occlusions (red box).} \label{fig:illustra}
\end{figure}

To address these challenges, we propose \textbf{Endo-TTAP}, a robust framework that synergizes hybrid supervision and multi-facet feature learning. As shown in Fig.\ref{fig:main}, our method advances endoscopic point tracking through three key designs. First, we introduce a \textbf{Multi-Facet Guided Attention} (MFGA) module that integrates multi-scale optical flow dynamics, semantic embeddings from DINOv2~\cite{oquab2023dinov2}, and explicit motion patterns to jointly predict point positions, occlusion states, and tracking confidence. Unlike prior works that treat occlusion and uncertainty as post-hoc corrections, MFGA enables lightweight Uncertainty and Occlusion Heads to directly model these factors during feature aggregation, enhancing robustness to endoscopic artifacts.
Second, we propose a two-stage training strategy with an \textbf{Auxiliary Curriculum Adapter} (ACA). In Stage I, synthetic data with perfect optical flow annotations warms up the model through uncertainty-occlusion regularization and flow consistency constraints. In Stage II, we bridge the natural-to-endoscopic domain gap via unsupervised flow distillation and semi-supervised pseudo-label learning, leveraging off-the-shelf trackers to generate reliable pseudo-labels for unlabeled frames. These pseudo labels are further refined through forward feature matching and backward filtering. 
Third, we validate our framework on a newly curated Endo-TAPC5 dataset, which spans five clinically challenging scenarios, complementing evaluations on established benchmarks (SurgT~\cite{allan2021stereo}, STIR~\cite{schmidt2024surgical}). Experiments demonstrate that our hybrid strategy outperforms both fully supervised and unsupervised baselines, particularly in the case of occlusions and long videos, as demonstrated in Fig.~\ref{fig:illustra}.


\section{Methodology}



\subsection{Two-stage Training of the Uncertainty and Occlusion Heads}
Endo-TTAP is designed for robust endoscopic tissue tracking through a hybrid-supervised framework that synergizes two-stage natural-to-real progressive adaptation and multi-facet feature fusion, as illustrated in Fig.~\ref{fig:main}. To achieve this, inspired by MFT~\cite{neoral2024mft}, we integrate two lightweight heads, i.e., the Uncertainty Head and the Occlusion Head (UO-Head), to enhance long-term tracking confidence and robustness. These two heads utilize a network structure with three alternating Conv2d and ReLU layers and are trained through a two-stage process for effective initialization and fine-tuning.

\noindent \textbf{Stage I - Supervised Learning with Synthetic Optical Flow.}
In Stage I, we leverage a pre-trained optical flow backbone with frozen weights to perform supervised learning on synthetic datasets with perfect optical flow annotations. To enable efficient warm-up, we introduce the \textbf{Auxiliary Curriculum Adapter} (ACA), which acts as a side branch to the frozen flow head. The ACA gradually increases the coefficient $\alpha$, controlling the ratio of the predicted optical flow used for concatenation, ensuring progressive training and smoother transitions for fine-tuning on endoscopic data.
The training is guided by three loss functions from supervised Optical Flow ($\mathcal{L}_{flow}^{1}$), Occlusion ($\mathcal{L}_{occ}^1$), and Uncertainty ($\mathcal{L}_{unc}^{1}$). Specifically, we refer to RAFT~\cite{teed2020raft} for $\mathcal{L}_{flow}^{1}$ and MFT~\cite{neoral2024mft} for $\mathcal{L}_{occ}^1$ which employs Binary Cross Entropy (BCE)~\cite{ruby2020binary}. The uncertainty loss $\mathcal{L}_{unc}^{1}$ can be expressed by the variance of the optical flow as follows:
\begin{equation}
\mathcal{L}_{unc}^{1}=\frac{1}{2 \sigma^{2}} \mathcal{L}_{H}\left(\left\|{x}-{x}^{*}\right\|_{2}\right)+\frac{1}{2} \log \left(\sigma^{2}\right),
\end{equation}
where ${x}$ is the predicted optical flow, ${x}^{*}$ is the optical flow ground truth, $\mathcal{L}_{H}$ is the Huber Loss~\cite{huber1992robust}, and $log(\sigma^{2})$ is the predicted uncertainty. 
The total loss for Stage I thus is \( L_{total}^{1}=\epsilon_{1}\mathcal{L}_{occ}^{1}+\mathcal{L}_{unc}^{1}+\mathcal{L}_{flow}^{1} \).


\noindent \textbf{Stage II - Holistic Learning with Flow and Pseudo Points.}
In Stage II, we fine-tune the UO-Head on real endoscopic datasets (SurgT~\cite{cartucho2024surgt} and STIR~\cite{schmidt2024surgical}) using a hybrid strategy comprising unsupervised optical flow training and semi-supervised pseudo-label learning.
For the first part, inspired by ARFlow~\cite{liu2020learning}, we introduce unsupervised optical flow training with real-world endoscopic data. Two classical loss functions, i.e., photometric loss $\mathcal L_{ph}^{2}$ and smoothness loss $\mathcal L_{sm}^{2}$ are utilized to optimize the forward and backward optical flows between two consecutive frames. We can thus simply formulate this unsupervised flow estimation loss in Stage II as $\mathcal L_{uflow}^{2}=\mathcal L_{ph}^{2}+\mathcal L_{sm}^{2}$.


Regarding the second part, to efficiently utilize a large number of unlabeled frames in the STIR~\cite{schmidt2024surgical} dataset, which only labels the first and last frame for each video sequence, we devise a Pseudo Label Generator (PLG) leveraging the powerful capability of off-the-shelf point tracking models, i.e., MFT~\cite{neoral2024mft} and CoTrackerV3~\cite{karaev2024cotracker3} as the teacher models. As demonstrated in Fig.~\ref{fig:main} (c), we first employ a feature-matching model XFeat~\cite{potje2024xfeat} to establish robust correspondence between the first frame $I_1$ and last frame $I_T$ of the sparsely labeled video. Then we select the key points with a threshold of 0.85, leaving about $N=6$ anchor pseudo point labels with the most essential features on $I_1$ and $I_T$. Then the teacher models take the $I_1$ and propagate forward through the entire video and generate pseudo points for all the remaining frames. On $I_T$, we compare the generated pseudo points with the anchor points and filter out those with an Euclidean distance larger than a circular radius $D=5$ pixels. Then we trace back and remove the corresponding unreliable pseudo points in other frames, and the final number of pseudo point labels is filtered from $N$ to $M$. Through such forward propagation and backward refinement, we manage to generate high-quality pseudo point labels for the unlabeled frames and facilitate semi-supervised training. Specifically, we introduce a point tracking loss $\mathcal{L}_{track}^2$ consisting of two parts, one for pseudo point labels supervision on the unlabeled frames and the other one for the last frame with ground truth point labels. Let $\mathcal{P}_{t}^m$ and $\mathcal{P}_{t}^{m*}
$ denote the predicted and pseudo-labeled $m^{th}$ points on the $t^{th}$ frame, and $\mathcal{P}_{T}^m$ and $\mathcal{P}_{T}^{m*}$ the predicted and ground-truth $m^{th}$ points on the last frame, we can formularize the tracking loss as:
\begin{equation}
{\mathcal{L}}_{track}^{2}= {\frac{\omega}{T-2}\sum_{t=2}^{T-2}\frac{1}{M^t}\sum_{m=1}^{M^t}\mathcal{L}_{H}({\mathcal{P}}_{t}^{m},{\mathcal{P}}_{t}^{m*})}+{\frac{1}{M^T}\sum_{m=1}^{M^T}\mathcal{L}_{H}({\mathcal{P}}_{T}^{m},{\mathcal{P}}_{T}^{m*})},
\end{equation}
where $M^t$ and $M^T$ is the number of tracked points in the frame $t$ and the last frame. $\omega$ is a weighting coefficient balancing the two loss terms and empirically set as 0.6 to put more weight on the supervision from the ground truth. Besides ${\mathcal{L}}_{track}^{2}$, we introduce the consistency loss ${\mathcal{L}}_{cons}^{2}$ which quantifies trajectory consistency by measuring the deviation of predicted points within a radius of $D=8$ pixels from their pseudo-labels. The consistency loss is defined as:
\begin{equation}
    \mathcal{L}_{cons}^{2}=\frac{1}{T-2}\sum_{t=1}^{T-2}\frac{1}{M^t}\sum_{m=1}^{M^t}\operatorname{CE}\left(\sigma(\mathcal{U}_{t}^{m}),\mathbb{1}\left[\|\mathcal{P}_{t}^{m}-\mathcal{P}_{t}^{m*}\|_2<D\right]\right),
\end{equation}
where $\sigma(\mathcal{U}_{t}^{m})$ represents the predicted uncertainty of the $m^{th}$ point of $t^{th}$ frame from the uncertainty head.
Moreover, we adopt an occlusion loss ${\mathcal{L}}_{occ}^{2}$ using Cross Entropy to enhance point occlusion prediction, which is a binary classification problem. 
Finally, the point supervised loss $\mathcal{L}_{\mathrm{point}}^{2}$ for fine-tuning the UO-Head is a weighted sum of the aforementioned losses with $\epsilon_2$ and $\epsilon_3$ as the balancing terms:
\begin{equation}    \mathcal{L}_{point}^{2}=\epsilon_{2}\mathcal{L}_{track}^{2}+\epsilon_{3}\mathcal{L}_{occ}^{2}+\mathcal{L}_{cons}^{2}.
\end{equation}
Totally, the holistic loss for Stage II is $\mathcal{L}_{total}^2=\mathcal{L}_{uflow}^{2}+\mathcal{L}_{point}^{2}$.

\subsection{Multi-Facet Guided Attention}
Effective uncertainty and occlusion estimation in endoscopic scenes requires robust integration of optical flow dynamics, visual-semantic cues, and multi-scale contextual features. Endoscopic imaging poses unique challenges: homogeneous tissue textures limit discriminative feature learning, while simple channel-wise features concatenation with varying dimensions (e.g., optical flow, image embeddings, hidden states) leads to suboptimal fusion. To address this, as illustrated in Fig.~\ref{fig:main} (d), we propose the Multi-Facet Guided Attention (MFGA) module, which aggregates six distinct feature types through a guided attention mechanism: (1) Forward and backward optical flow outputs; (2) Four middle features from the flow backbone, including Correlation Cost-Volume, Hidden GRU State, Context Features, and Motion Features; and (3) DINOv2~\cite{oquab2023dinov2} semantic embeddings.


\noindent \textbf{Feature Alignment and Guided Fusion.} First, the dual optical flow outputs are concatenated along the channel dimension to produce $F_{flow}$. Both $F_{flow}$ and the middle features $F_{m}^{i}$ $(i=1,2,3,4)$ undergo independent channel attention computation. The optical flow features are then fused with the current image feature extracted by DINOv2~\cite{oquab2023dinov2} to obtain a hybrid feature $F_{hybrid}$ of visual and optical flow. Then, to ensure compatibility for cross-modal fusion, we map all features to a shared latent space ($C_{ls} = 128$) through linear layers, \(F_{hybrid}^c \in \mathbb{R}^{H \times W \times C_{ls}}\) and \(F^{i,c}_m \in \mathbb{R}^{H \times W \times C_{ls}}\). The fusion process employs a query-key-value attention mechanism where \(F_{hybrid}^c\) serves as the Query and Middle features \(F^{i,c}_m\) act as Key-Value.
The guided attention score $F_{GA}$ for each \(i^{th}\) middle feature is computed as:
\begin{equation}
    F_{GA} = \sum_{i=1}^4 \left[\text{Softmax}\left(\frac{F_{hybrid}^c \cdot (F^{i,c}_m)^\top}{\sqrt{C_{ls}}}\right) \cdot F^{i,c}_m\right],
\end{equation}
with \(\text{Attn}_i \in \mathbb{R}^{H \times W \times H \times W}\) capturing spatially adaptive correlations between hybrid and middle features, and \(\sqrt{C_{ls}}\) ensures stable gradient propagation.
By explicitly modeling feature correlations through guided attention, the MFGA module resolves ambiguities arising from tissue homogeneity and enhances robustness under occlusion and deformation.



\section{Experiments and Results}
\subsection{Experimental Settings}
We use different datasets for training at different stages. In Stage I, we use the Sintel~\cite{butler2012naturalistic} and FlyThing3D~\cite{mayer2016large} datasets for supervised optical flow training. During Stage II, we transition to the training datasets of SurgT~\cite{cartucho2024surgt} from MICCAI 2022 Challenge with 125 videos and STIR~\cite{schmidt2024surgical} from MICCAI 2024 Challenge with 576 videos for unsupervised optical flow training and semi-supervised training with pseudo-labeled points. 
For testing, we adhere to the settings of Ada-tracker~\cite{guo2024ada} to assess performance on the SurgT~\cite{cartucho2024surgt} dataset. This includes standard 2D metrics—accuracy, error, and robustness—and 3D metrics—error and robustness. Similarly, we follow the official STIR~\cite{schmidt2024surgical} Challenge protocols, utilizing 2D metrics such as accuracy and end point error. Additionally, the Endo-TAPC5 dataset combines the Hamlyn~\cite{ye2017self}, Cholec80~\cite{rios2023cholec80}, and a private collection of 40 surgical videos from XYZ Hospital, featuring five key challenges: \textit{Tissue Deformation}, \textit{Instrument Occlusion}, \textit{Jitter}, \textit{Reflection}, and \textit{Smoke}. This dataset follows the same rigorous labeling and evaluation standards as STIR~\cite{schmidt2024surgical}.

Our Endo-TTAP framework is built upon SEA-RAFT~\cite{wang2024sea}, utilizing its optical flow backbone and flow head, both of which remain frozen throughout the training process. We adopt the same optimizer and data augmentation strategies as in SEA-RAFT~\cite{wang2024sea}. In the warm-up stage I, we performed supervised training for 10,000 iterations with a learning rate of 2e-5 and a batch size of 8. The Auxiliary Curriculum Adapter module is configured to map $\alpha$ exponentially from 1e-5 to 0.3. For Stage II, we alternate between semi-supervised point-based training and unsupervised optical flow training for 2,500 iterations each, resulting in a total of 10,000 training iterations. In this stage, $\alpha$ is exponentially mapped from 0.3 to 1.0, with learning rates of 1e-6 and 1e-5 for the two phases, respectively. The loss balancing coefficients are empirically set to $\epsilon_1=\epsilon_3=10$ and $\epsilon_2=1.2$. For evaluation, we compare our approach with original challenge submissions and also with several recent methods~\cite{guo2024ada,lukezic2017discriminative,karaev2024cotracker3,neoral2024mft,serych2024mftiq}. Results for challenge submissions are obtained from their respective challenge reports.
All training experiments are performed on two NVIDIA RTX 4090 GPUs.

\begin{table}[t]
\centering
\caption{\textbf{Quantitative comparison of the tissue tracking performance.} Our approach consistently outperforms other baselines on three datasets. The best results are bolded. Runner-up results are underlined. "-" indicates infeasible results or reproduction.} 
\resizebox{0.95\textwidth}{!}{ 
\label{tab:table} 
\begin{tabular}{l|ccccc|cc|cc}
\bottomrule
\textbf{Dataset} & \multicolumn{5}{c|}{\textbf{SurgT~\cite{cartucho2024surgt}}}                                                    & \multicolumn{2}{c|}{\textbf{STIR~\cite{schmidt2024surgical}}} & \multicolumn{2}{c}{\textbf{Endo-TAPC5}} \\ \hline
Method           & Rob2D$\uparrow$         & Acc2D$\uparrow$         & Err2D$\downarrow$           & Rob3D$\uparrow$         & Err3D$\downarrow$           & Acc2D$\uparrow$          & Epe2D$\downarrow$           & Acc2D$\uparrow$             & Epe2D$\downarrow$             \\ \hline
Jmees            & 0.868          & 0.818          & 5.3$\pm$2.4          & 0.878          & 2.7$\pm$1.9          & 0.690            & -                & -                  & -                  \\
SRV              & 0.476          & 0.681          & 15.4$\pm$7.5         & 0.71           & 8.5$\pm$13.7         & 0.344           & -                & -                  & -                  \\
MEDCVR           & 0.702          & 0.509          & 7.9$\pm$5.8          & 0.832          & 9.2$\pm$39.7         & -               & -                & -                  & -                  \\
ETRI             & 0.802          & 0.693          & 12.1$\pm$7.4         & 0.909          & 5.7$\pm$5.2          & -               & -                & -                  & -                  \\
RIWOlink         & 0.807          & 0.737          & 8.0$\pm$4.5          & 0.894          & 5.8$\pm$9.2          & -               & -                & -                  & -                  \\
ICVS-2Ai         & 0.872          & 0.816          & 6.7$\pm$3.5          & 0.901          & 2.6$\pm$2.3          & 0.664           & -                & -                  & -                  \\
Ada-tracker~\cite{guo2024ada}      & 0.894          & 0.833          & 5.2$\pm$3.3          & 0.912          & 2.1$\pm$2.0          & -               & -                & -                  & -                  \\ \hline
CSRT~\cite{lukezic2017discriminative}             & 0.872          & 0.769          & 5.7$\pm$2.6          & 0.894          & 3.3$\pm$3.8          & 0.586           & 44.403           & 0.420                & 132.919            \\
CotrackerV3~\cite{karaev2024cotracker3}      & 0.876          & 0.831          & 5.6$\pm$1.2          & 0.887          & 3.2$\pm$2.3          & 0.745           & 22.514           & 0.585              & 94.662             \\
MFT~\cite{neoral2024mft}              &\uline{0.923}          & 0.902          & 3.1$\pm$1.5          & 0.932          & 1.4$\pm$1.9          & \uline{0.776}           & 16.317           &\uline{0.706}            & \uline{20.186}             \\
MFTIQ~\cite{serych2024mftiq}    & 0.913           & \uline{0.917}          & \uline{2.1$\pm$0.8}          & \uline{0.940}           & \uline{1.0$\pm$1.4}          & 0.768           & \uline{15.659}           &  {0.686}   & 20.563                   \\
Endo-TTAP(Ours)   & \textbf{0.931} & \textbf{0.922} & \textbf{1.9$\pm$1.0} & \textbf{0.946} & \textbf{0.9$\pm$0.8} & \textbf{0.784}  & \textbf{13.324}  & \textbf{0.742}    & \textbf{17.831}        \\ \bottomrule
\end{tabular}
}
\end{table}

\begin{figure}[t]
\centering
\includegraphics[width=0.95\textwidth]{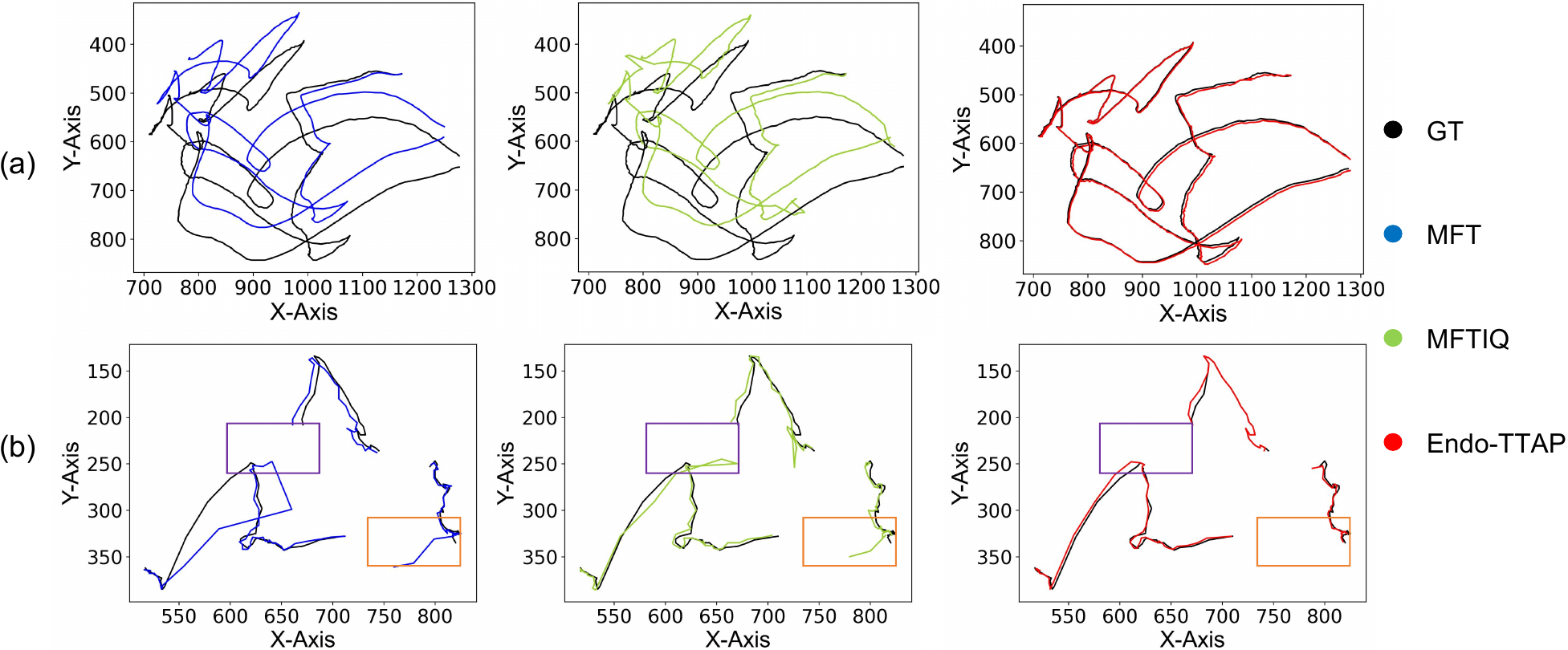}
\caption{Qualitative comparison of the tracking trajectory by MFT~\cite{neoral2024mft}, MFTIQ~\cite{serych2024mftiq} and our Endo-TTAP for (a) normal and (b) challenging cases. The rectangle boxes highlight more accurate and stable point tracking of our approach.} \label{fig:trajectory}
\end{figure}

\subsection{Results and Analysis}

As shown in Table.~\ref{tab:table}, our Endo-TTAP achives state-of-the-art performance across all metrics and datasets. Specifically, on the SurgT~\cite{cartucho2024surgt} dataset, our method significantly outperforms the Challenge submissions, demonstrating higher 2D/3D accuracy and lower 2D/3D error. Notably, our approach achieves superior robustness scores, highlighting its effectiveness in complex endoscopic scenarios. On the STIR~\cite{schmidt2024surgical} dataset, our method achieves an endpoint error (EPE) of 13.324, a substantial improvement over the second-best result of 15.659. This underscores its capability in long-term video tracking, as also illustrated in Fig.~\ref{fig:trajectory} (a), which visualizes the stable point trajectory following over an extended video sequence. 
On our curated Endo-TAPC5 dataset, which encompasses diverse challenges in tissue point tracking, our method outperforms the runner-up, with metrics of 0.036 for Acc2D and 2.355 for Epe2D. This demonstrates its ability to maintain high tracking accuracy and robustness under varying conditions. 
As highlighted in Fig.~\ref{fig:trajectory} (b), the purple box shows rapid endoscope movement, where MFT~\cite{neoral2024mft} and MFTIQ~\cite{serych2024mftiq} exhibit significant tracking drift, while Endo-TTAP remains stable.
The orange box illustrates a point reappearing after surgical instrument occlusion. While MFT~\cite{neoral2024mft} and MFTIQ~\cite{serych2024mftiq} struggle to re-establish tracking, Endo-TTAP accurately recovers the point trajectory. 
We strongly recommend reviewing our supplementary videos for a more comprehensive and intuitive visualization of the tracking performance.


\begin{table}[t]
\centering
\setlength{\tabcolsep}{4pt} 
\begin{minipage}[t]{0.48\textwidth}
\centering
\caption{Ablation studies with different losses at different stages.}
\label{tab:abl_loss}
\resizebox{\linewidth}{!}{ 
\begin{tabular}{clcc|cc|cc}
\bottomrule
\multicolumn{4}{c|}{\textbf{Loss function}}      & \multicolumn{2}{c|}{\textbf{STIR~\cite{schmidt2024surgical}}} & \multicolumn{2}{c}{\textbf{Endo-TAPC5}} \\ \hline
\multicolumn{2}{c}{$\mathcal{L}_{total}^{1}$} & $\mathcal{L}_{uflow}^{2}$     & $\mathcal{L}_{point}^{2}$   & Acc2D$\uparrow$           & Epe2D$\downarrow$            & Acc2D$\uparrow$             & Epe2D$\downarrow$            \\ \hline
\multicolumn{2}{c}{$\checkmark$}    & \ding{55}        & \ding{55}     & 0.723            & 22.327          & 0.605              & 54.398            \\
\multicolumn{2}{c}{$\checkmark$}    & {$\checkmark$}        & \ding{55}      & 0.748            & 19.982          & 0.637              & 43.632            \\
\multicolumn{2}{c}{$\checkmark$}    & \ding{55}        & {$\checkmark$}      & 0.776            & 15.754          & 0.708              & 26.438                    \\
\multicolumn{2}{c}{$\checkmark$}    & {$\checkmark$}        & {$\checkmark$}      & \textbf{0.784}            & \textbf{13.324}          & \textbf{0.742}             & \textbf{17.831}            \\ \bottomrule
\end{tabular}
}
\end{minipage}
\hspace{0.02\textwidth} 
\begin{minipage}[t]{0.48\textwidth}
\centering
\caption{Ablation studies of MFGA and DINOv2~\cite{oquab2023dinov2} visual embedding.}
\label{tab:abl_mfga}
\resizebox{\linewidth}{!}{ 
\begin{tabular}{cc|cc|cc}
\bottomrule
\multicolumn{2}{c|}{\textbf{Module}} & \multicolumn{2}{c|}{\textbf{STIR~\cite{schmidt2024surgical}}} & \multicolumn{2}{c}{\textbf{Endo-TAPC5}} \\ \hline
MFGA         & DINOv2~\cite{oquab2023dinov2}        & Acc2D↑      & Epe2D↓      & Acc2D↑         & Epe2D↓        \\ \hline
\ding{55}            & \ding{55}             & 0.738       & 16.609      & 0.683          & 19.190        \\
{$\checkmark$}            & \ding{55}             & 0.746       & 16.081      & 0.702          & 18.563        \\     
{$\checkmark$}            & {$\checkmark$}             & \textbf{0.784}       & \textbf{13.324}      & \textbf{0.784}          & \textbf{17.832}        \\
\bottomrule
\end{tabular}
}
\end{minipage}
\end{table}



\noindent \textbf{Ablation Study.}
We conduct ablation studies to validate key design choices in the Endo-TTAP framework. Firstly, as shown in Table~\ref{tab:abl_loss}, we analyze the loss contributions in our two-stage training. We can find that the unsupervised optical flow learning loss $\mathcal{L}_{uflow}^{2}$ and the semi-supervised pseudo point tracking loss $\mathcal{L}_{point}^{2}$ both help improve the tracking results on STIR~\cite{schmidt2024surgical} and our Endo-TAPC5 datasets. Their combined use yields optimal results, confirming the effectiveness of our hybrid two-stage training strategy. Besides, we investigate our Multi-Facet Guided Attention (MFGA) module and the integration of visual features from DINOv2~\cite{oquab2023dinov2}. As evidenced by Table~\ref{tab:abl_mfga}, the MFGA module improves the tissue point tracking performance on both datasets and is further improved when additional visual features from DINOv2~\cite{oquab2023dinov2} are introduced.

\section{Conclusion}

In this work, we introduce \textbf{Endo-TTAP}, a robust framework for endoscopic tissue tracking that addresses challenges like sparse annotations, complex deformations, and instrument occlusion. Our method integrates three innovations: (1) the Multi-Facet Guided Attention (MFGA) module, which combines multi-scale flow features, semantic embeddings, and motion patterns to predict point positions, occlusion states, and tracking confidence; (2) a two-stage training strategy with an Auxiliary Curriculum Adapter (ACA), enabling progressive initialization and fine-tuning under hybrid flow and pseudo-point supervision; and (3) a hybrid supervision framework that leverages unsupervised flow distillation and semi-supervised pseudo-label learning to reduce dependence on sparse annotations.
Extensive experiments on the SurgT~\cite{cartucho2024surgt}, STIR~\cite{schmidt2024surgical}, and our Endo-TAPC5 datasets demonstrate that Endo-TTAP achieves state-of-the-art performance, excelling in long-term tracking and challenging conditions such as occlusion and tissue deformation. Ablation studies confirm the contributions of each component, including the MFGA module and DINOv2~\cite{oquab2023dinov2} embeddings.
Endo-TTAP advances endoscopic tissue tracking, offering a practical solution for surgical navigation and scene understanding. Future work will explore tracking acceleration and multi-point co-tracking for broader application scenarios. 

\bibliographystyle{splncs04}
\bibliography{MICCAI2025}
%




\end{document}